\documentclass[sigconf, nonacm]{acmart}
\usepackage{amsmath}
\DeclareMathOperator*{\argmax}{argmax}
\usepackage[ruled,vlined,linesnumbered]{algorithm2e}
\SetKw{Continue}{continue}
\SetKw{Break}{break}
\SetKw{Return}{return}
\usepackage{multirow}
\usepackage{framed}
\usepackage{subcaption}
\newtheorem{defn}{Definition}

\usepackage{todonotes} 

\begin{document}

\title{Towards Repairing Neural Networks Correctly}
 \author{Guoliang Dong}
  \affiliation{%
    \institution{Zhejiang University}
  }
  \email{dgl-prc@zju.edu.cn}

  \author{Jun Sun}
  \affiliation{
    \institution{Singapore Management University}
  }
  \email{junsun@smu.edu.sg}

 \author{Jingyi Wang}
  \affiliation{
    \institution{Zhejiang University}
  }
  \email{wangjyee@zju.edu.cn}

\author{XingenWang}
  \affiliation{%
    \institution{Zhejiang University}
  }
  \email{newroot@zju.edu.cn}

  \author{Ting Dai}
  \affiliation{%
    \institution{Huawei International Pte Ltd}
    }
  \email{daiting2@huawei.com}
  
    \author{Xinyu Wang}
  \affiliation{%
    \institution{Zhejiang University}
  }
  \email{wangxinyu@zju.edu.cn}
  
\begin{abstract}
Neural networks are increasingly applied to support decision making in
safety-critical applications (like autonomous cars, unmanned aerial vehicles and
face recognition based authentication). While many impressive static
verification techniques have been proposed to tackle the correctness problem of
neural networks, existing static verification techniques are still not as
scalable as hoped. Furthermore, it is possible that static verification may
never be sufficiently scalable to handle real-world neural networks. In this
work, we propose a runtime repairing method to ensure the correctness of neural
networks. Given a neural network and a safety property, we first adopt
state-of-the-art static verification techniques to verify the neural networks.
In the case that the verification fails, we strategically identify locations to
introduce additional gates which ``correct'' neural network behaviors at runtime
whilst keeping the modifications minimal. Experiment results show that our
approach effectively generates neural networks which are guaranteed to satisfy
the properties, whilst being consistent with the original neural network most of
the time. 
\end{abstract}
\maketitle

\section{Introduction}

\begin{figure*}[t]
\centering
\includegraphics[width=0.8\textwidth]{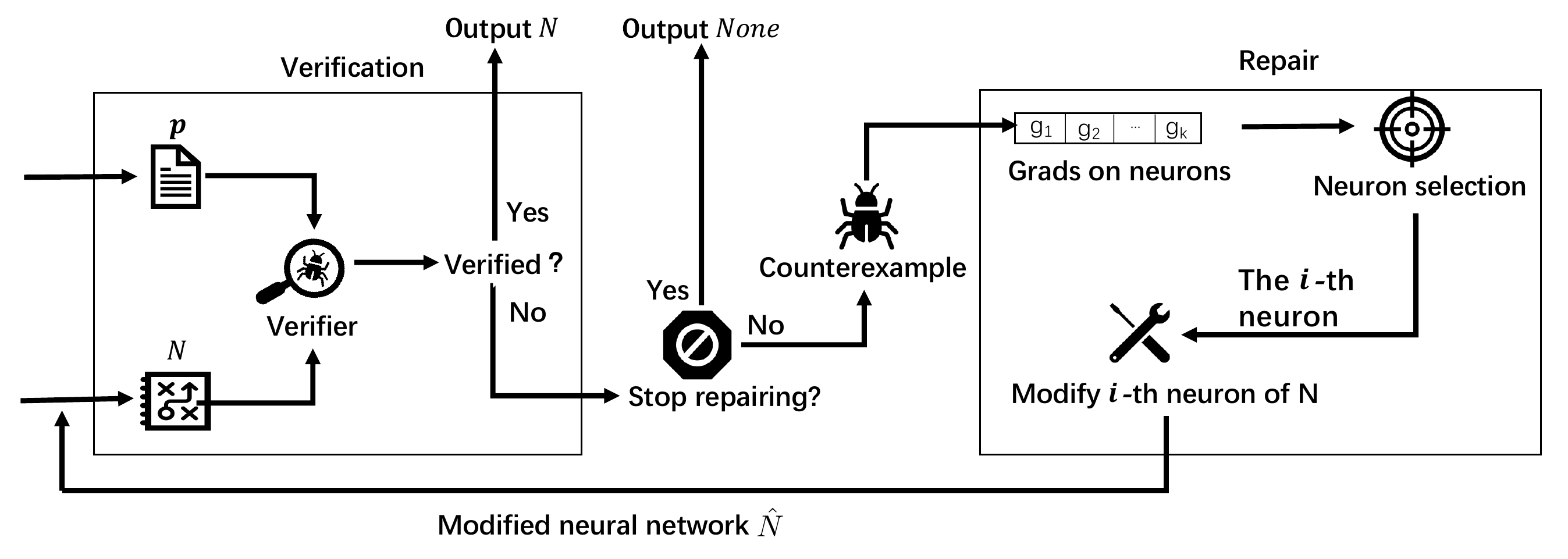}
\caption{Overall framework}
\label{fig:frame}
\end{figure*}

Deep neural networks (DNNs) are widely applied to a variety of applications thanks to their exceptional performance, such as facial
recognition~\cite{schroff2015facenet}, sentiment analysis~\cite{tang2015document}, and malware detection~\cite{yuan2014droid}. In
addition, they are increasingly applied in safety-critical systems such as medical diagnosis~\cite{vieira2017using}, self-driving
cars~\cite{bojarski2016end} and aircraft collision avoidance for unmanned aircraft (ACAS Xu)~\cite{acasxu}, which highlights the
growing importance of DNNs' safety and reliability. DNNs are, however, known to be brittle to attacks such as adversarial
perturbations~\cite{goodfellow2014explaining}. That is, a slight perturbation on an input can cause a DNN to make a decision in an
unexpected and incorrect way. Worse yet, due to the black-box nature of DNNs, it is extremely hard to `debug' and repair such erroneous
behaviors. 

Existing efforts on ensuring that DNNs behave correctly roughly fall into two
categories. One is on static verification of DNNs, including efforts such as
Reluplex~\cite{reluplex}, MIPVerify~\cite{MIPVerify} and
DeepPoly~\cite{deeppoly}. This group of works focus on statically verifying a
specific property of interest, e.g., a reachability condition or local
robustness. The aim is to provide a formal guarantee on the correctness if the
property is satisfied. This group of works suffer from two limitations. First,
these works are not yet scalable enough to handle certain real-world DNNs (which may
contain thousands or even millions of neurons). Although more and more
sophisticated verification algorithms have been
developed~\cite{refinepoly,neurify}, it is entirely possible that they may never
be scalable enough. Second, these approaches do not answer the natural question:
what the subsequent measure that one should take if the DNN is not verified?
While we can typically construct a counterexample and then fix the `bug' in the
setting of program verification, how to improve a DNN with the verification
result is far more complicated. Note that simply discarding the problematic DNN
and training a new one from scratch is not always feasible due to the high time
and computing costs. Worst yet, there is no guarantee that the newly trained
neural network is safe either.

The second group of works focus on improving DNNs by repairing them~\cite{ma2018mode,sohn}. 
One idea is to retrain a target model with adversarial samples which can be
generated using attacking tools~\cite{ma2018mode, li2016general}. Although the
retrained models typically have improved robustness against adversarial perturbations, they do
not guarantee that the retrained model is correct. Indeed, it has been shown
that such retrained models are often subject to further adaptive adversarial
attacks~\cite{tramer2017ensemble}. Another line of work on repairing
DNNs~\cite{sohn, mm} is to, given a DNN, modify the weights of neurons which
contribute to certain specific behaviours. For instance, Arachne, proposed
in~\cite{sohn}, modifies the weights of certain neurons guided by a fitness
function to prune misbehaviours caused by certain inputs. However, Arachne
similarly does not guarantee the correctness of the repaired model. Goldberger \emph{et al.} proposed in~\cite{mm} a
verification-based approach to modify neural weights of the output layer such
that the DNN satisfies a given property. Different from the previous repairing
approaches, this approach provides a formal guarantee on the correctness of the
repaired model based on verification techniques. However, their repair is
limited to the output layer, which is shown to have limited capability in
handling DNNs trained for realistic safety-critical systems such as ACAS Xu.
Further, their approach is limited to correcting the DNN’s behavior on one
concrete input, which has limited usefulness as there could be many such inputs. Repairing the DNN
so that it always behaves correctly with respect to a desirable property is
both more useful and technically challenging.

In this work, we propose a verification-based approach for repairing DNNs.
The goal of our work is to repair a DNN such that it is guaranteed to satisfy user-provided properties. 
Different from~\cite{mm}, our approach is not restricted to modifying the output
layer (which is often too late). Rather, we propose to identify and modify the
most relevant neurons for violating the desired property through the guidance of
the verification results. We further incorporate abstraction refinement techniques to minimize
the modification so that the modification is only relevant in limited regions of
input space where correctness cannot be verified. 

Figure~\ref{fig:frame} presents the overall workflow of our approach. There are
two main parts, i.e., network verification (on the left) and network repair
(on the right). Given a network $N$ and a user-provided property $\phi$, we first
check whether network $N$ satisfies $\phi$ or not with a static verification
engine. If the property is violated and the termination condition, e.g., timeout, has not been met, 
we identify a counterexample. Afterwards, we first compute the gradient of each neuron with respect to the violation loss of the counterexample, and then select the neuron which is most
‘responsible’ for the violation according to the magnitude of the gradients. Intuitively, the magnitude
of the gradients measures the contribution of the neurons to DNN prediction
result~\cite{jsma}. The gradients of the network's output with respect to the
neurons thus can be regarded as a measure on the neurons' contribution on
violating the property. We then tune the weights of the selected neuron and
obtain a modified model $\hat{N}$. Iteratively, we take $\hat{N}$ as the new input model
and repeat the above process until the network is verified or the termination
condition is met.  


We have implemented our approach as a self-contained prototype based on the
DeepPoly verification engine and evaluate it on 41 models for two kinds of
tasks, i.e., aircraft collision avoidance (ACAS Xu) and image classification
(MNIST and CIFAR10). For ACAS Xu, we apply our approach to the models which
violate at least one of the properties. For the image classification, we focus
on the robustness property, i.e., we apply our approach to repair the models so
that they are robust against perturbation which are limited to certain region of
the input images. In total, we have a set of 587 repair tasks. Overall, our
approach achieves 98.46\%, 94.77\% and 96.67\% success rate on models of ACAS Xu,
MNIST and CIFAR10 respectively. On average, the time overhead to repair a target
model is 6.5, 2.7 and 2.07 minutes for models of ACAS Xu, MNIST, and CIFAR10
respectively. To show that the repaired DNNs are faithful to the original DNN, we
measure the fidelity of each repaired DNN, and the results show that the
repaired models have a high fidelity of 97.51\% with respect to the
original DNNs.

In summary, we make the following technical contributions. 
\begin{itemize}
    \item We propose an effective and efficient verification-based framework for
    repairing DNNs. Different from existing approaches, the repaired DNN is guaranteed to satisfy the property.
    \item We propose to repair DNNs based on adjusting weights of neurons which are most responsible for violating the property, through an optimization algorithm.
    \item We implement a self-contained toolkit for repairing DNNs and evaluate our approach on two tasks over three datasets. The results
    show our approach can effectively repair DNNs.
\end{itemize}

The remainder of the paper is organized as follows. We review relevant background in Section 2, and present our approach in detail in Section 3. We evaluate our approach 
and discuss the experiment results in Section 4.  We review related work in Section 5 and conclude in Section 6.

\section{Background}
\label{sec:bg}
In this section, we briefly review relevant background. \\

\noindent \emph{Deep neural networks} In this work, we focus on feedforward neural networks (FNNs) for various classification tasks. We remark that in theory our approach can be extended to support other kinds of DNNs as long as static verification techniques for those networks are available. A FNN consists of an input layer, multiple hidden layers, and an output layer.  Let $N$ be an FNN. We denote $N = f_0 \circ f_1 \circ \cdots \circ f_l$ where $f_0$ is the input layer, $f_l$ is the output layer, and $f_i$ ($i\in[1, l-1]$) in between is the hidden layer. N can be regarded as a function $N: X \rightarrow Y$ mapping an input $x\in X$ to a label $c \in  Y$. Given an input $x\in X$, for each layer $f$, it computes the output as follows.
\begin{align}
& f_0 = x\\
& f_i  = \sigma(W^if_{i-1}+B^i)
\end{align}
where $W^i$ and $B^i$ are the weights and biases of the neurons in $i$-th layer respectively, and $\sigma$ is the activation
function such as maxout~\cite{maxout}, rectified linear unit (ReLU)~\cite{relu} and hyperbolic tangent (tanh). In this work, we
focus on ReLU which coverts any negative input to zero and keep the positive input unchanged. Note that here we use $f$ to denote
the layer of neurons or the output of layer $f$ depending on the context. The predicted label $c$ is obtained from the output
vector of the final layer $f_l$:  $c = \argmax_i(f_l^i)$ where $f_l^i$ denotes the $i$-th value in output $f_l$. \\

\begin{table}[t]
\caption{An example of the reachability property}
\begin{tabular}{@{}lll@{}}
    \toprule
    Input & $x=[x_1,x_2,x_3,x_4,x_5]$ &  \\
    Output & $y=[y_1,y_2,y_3,y_4,y_5]$ &  \\
    Input proposition   & $x_1 \geq 55947.691$, $x_4 \geq 1145$ and $x_5\leq 60$ &  \\
    Output proposition  & $y_1$ is not the maximal score & \\
    \bottomrule
   \end{tabular}
    \label{tab:exp1}
\end{table}

\noindent \emph{Properties and verification of neural networks} The problem of
verifying neural networks is to provide formal guarantees about if a given
network satisfies a certain property of interest. There are a variety of
properties addressing different concerns. In this work, we focus on a class of
reachability properties which are defined based on an input proposition $\phi$
and an output proposition $\omega$. Table~\ref{tab:exp1} exemplifies a property
for ACAS Xu models~\cite{acasxu}, which describes that if the intruder aircraft is
distant and is significantly slower than the ownship, the score of a ``Clear-of-Conflict''
advisory should not be the maximal score.
Formally, the problem is to check if the following assertion holds.
\begin{align}
\forall x \vDash \phi.~ N(x) \vDash \omega 
\label{fm:p}
\end{align}
Intuitively, the property states that if an input satisfies certain constraint $\phi$, the neural network output must satisfy
$\omega$. A verification algorithm may produce three results. One is that the property is verified. One is that the property is
violated and a counterexample is generated (i.e., one input $x$ such that $x \vDash \phi \land N(x) \neg \vDash \omega$). The last
one is that the algorithm fails to verify or falsify the property (e.g., timeout or outputs an `unknown' result), due to the
limitation of existing verification techniques. 

Among many existing verification toolkits for neural networks, we focus on \emph{DeepPoly} in this work. DeepPoly is a
state-of-the-art neural network verifier~\cite{deeppoly}. Verification with DeepPoly comprises two steps. Firstly,  it adopts
abstract interpretation to over-approximate the reachable set at each layer, starting with $\phi$ at the input layer. Through
layer-by-layer propagation, DeepPoly obtains an abstract representation of the reachable set for each class. Then, based on the
approximation of the output of each label, DeepPoly checks whether $\omega$ is satisfied. Let $\phi=(l, u)$ be the input
constraint where $l$ and $u$ is the lower bound and upper bound of the input respectively,  we then formalize the first step of
DeepPoly as follows.
\begin{align}
\mathcal{A}(N,\phi) = \{(a_1^{\leq}, a_1^{\geq}), \dots, (a_k^{\leq}, a_k^{\geq}) \}
\end{align}
where $a_i^{\leq}$ and $a_i^{\geq}$ are the lower bound and upper bound of the abstract domain of label $i$. During the second step,
DeepPoly checks whether $\{(a_1^{\leq}, a_1^{\geq}), \dots, (a_k^{\leq}, a_k^{\geq}) \}$ satisfies $\omega$.  For example, if $\omega$
requires that the score assigned to label 1 is always the largest among all the labels for any input $x \in \phi$, DeepPoly checks
whether the lower bound of label 1 is greater than the upper bound of any of the other labels. If it is the case, the property is verified; otherwise, DeepPoly reports that the verification fails.

%
%
%
%
%
%

\section{Our Approach} 
\label{sec:app} 
In this section, we present our approach in detail. The goal of our work is to repair a neural network such that the user-provided property is guaranteed to be satisfied. Our approach can be categorized as verification-based repair. The
overall idea is to identify a minimal set of input regions in which the property is not verified and then tune the weights of the
neurons which are `most' responsible for violating the property in the regions so that the property is satisfied. In
the following, we first define the repair problem, and then present the details of our solution.

\subsection{Problem definition}
\label{sec:pld}
 Now, we formally describe our repair problem as follows. 
\begin{defn}
\label{def:pld}
    Let $N$ denote a neural network; let $\phi$ denote the input constraint and
    let $\omega$ denote the output constraint. Assuming that $\phi$ consists of
    a set of disjoint partitions: $\phi=\{\phi_1,\phi_2,\dots,\phi_n\}$ (i.e., partitions of the input space), and $N$
    violates $\omega$ in some of the partitions. Our repair problem is to find a neural network $\hat{N}$ for each erroneous
    partition $\phi'$ such that $\forall x \vDash \phi',~ \hat{N}(x) \vDash \omega$.
\end{defn}

According to definition~\ref{def:pld}, our approach returns a set of repaired models and each of them is responsible for a certain
input subregion $\phi'$ in which the property is violated. That is, when we use the returned models to predict, we first locate the subregion where the input sample
belongs to, and then use the corresponding model to perform the prediction. Algorithm~\ref{alg:predict} shows at a high-level how to use
the repaired models, where $x$ is the input sample, $N$ is the original model and $R$ is the set of repaired models of $N$. As aforementioned in Section~\ref{sec:bg}, we focus on the reachability
properties. That is, given an input from a certain input region, any output returned
by a repaired model is acceptable as long as it satisfies the output constraint,
even the output is different from the one returned by the original model.

Note that compared to having a repaired model for all inputs, this way of
repairing the neural network allows more flexibility. First, a well trained and
tested neural network often behaves correctly in many of the regions and thus we
should avoid to repair the neural network in those regions. Second, repairs in
different regions could be different, which potentially allows us to utilize
characteristics which are specific to certain region to assist the repair.
Furthermore, we remark that our repair is not the DNN minimal modification
problem proposed in~\cite{mm}. More specifically, we do not put any ``quality"
constraint, e.g, a minimal distance between the repaired and the buggy network,
during repairing because our repair is property-oriented. That is, we assume
that the property is critical and must be satisfied (e.g., collision avoidance
for unmanned aircraft); and if a repaired model can satisfy a given property, we
say the repair is successful regardless of how many modifications are made. One
may concern that without such constraints on the repaired network, our approach
could return a new network that is repaired but behaves very differently from
the original network. We will address this concern later. 

\begin{algorithm}[t]
    \caption{$repair\_predict(x, N, R)$}
    \label{alg:predict}
    \For{$i=1$ to $size(R)$}{
      let $\hat{N}$ be the $i$-th repaired models in $R$\;
      let $\hat{\phi}$ be the input region corresponding to $\hat{N}$\;
      \If{$x \in \hat{\phi}$}{
        \Return $\hat{N}(x)$\;
      }
    }
    \Return $N(x)$
\end{algorithm}




\subsection{From Verification to Optimization}
Our overall algorithm is shown in Algorithm~\ref{alg:overall} which takes as
inputs the original model $N$, the input constraint $\phi$, the output
constraint $\omega$, a bound on the number of neurons to modify $\alpha$, the
maximum number of times a single neuron is allowed to be modified $\beta$ and
the step size $\eta$. First, we employ a verifier to verify $N$ against the
property. If the property is satisfied, we return $N$ without any modifications.
Otherwise, we check if the input domain $\phi$ can be further partitioned. While
in theory, it is always possible to partition the input domain, in practice,
existing verifiers often have restrictions on the form of input constraints
(e.g., a range constraint subject to certain further restriction in
DeepPoly~\cite{deeppoly}) and thus it is not always possible to partition. If
affirmative, we partition $\phi$ into two non-overlapping constraints $\phi_1$
and $\phi_2$ and repair $N$ for each partition separately. At line 8, we
assemble a neural network based on the repaired results $\hat{N}_1$ and
$\hat{N}_2$, i.e., by adding a gate which directs the input to $\hat{N}_1$ if
$\phi_1$ is satisfied or $\hat{N}_2$ otherwise. The partition strategy differs
according to the verifier engine. One simple and general (and importantly
efficient) strategy is bisection (i.e., by bisecting a range into two
equal-sized ranges). If $\phi$ cannot be further partitioned, a counterexample
$ct$ is generated and we invoke Algorithm~\ref{alg:repair} to repair $N$.  

\begin{algorithm}[t]
\caption{$overall(N, \phi, \omega, \alpha, \beta, \eta)$}
\label{alg:overall}
    verify $N$ with a verifier\; \If{$N$ is verified} {\Return{$N$}\;}
    \If{$\phi$ can be further partitioned} {
        partition $\phi$ into $\phi_1$ and $\phi_2$\; 
        let $\hat{N}_1$ be $overall(N, \phi_1,
        \omega, \alpha, \beta, \eta)$\; 
        let $\hat{N}_2$ be $overall(N, \phi_2, \omega, \alpha, \beta, \eta)$\; 
        \Return{an assemble of $\hat{N}_1$ and $\hat{N}_2$}\;
    } 
    \Else{
        \If{a counterexample $ct$ is generated} {
            \Return{$repair(N,\phi, \omega, ct,\alpha, \beta, \eta)$}\;
        }
    }
\end{algorithm}

Given a particular input region (i.e., $\phi$) in which the property is violated, Algorithm~\ref{alg:repair} aims to repair the
neural network by tuning the weights of the relevant neurons. The question is then: which neurons do we tune (so that the neuron
network is repaired by tuning a minimal set of neurons) and how do we tune the neurons? Our answer to the question is to solve an
optimization problem. That is, given the property and the counterexample $ct$, we define a loss function and then minimize the
violation loss of $ct$ by modifying the outputs of neurons.
Formally,
\begin{align}
\label{fm:repair}
\min_{\hat{N}} loss(ct, \hat{N})
\end{align}
where $\hat{N}$ is a repaired model satisfying the property. To reduce the search space, we restrict $\hat{N}$ such that $\hat{N}$
and $N$ have the same structure and weights except that some neurons $\hat{N}$ have a constant activation value. Note that the above loss function is defined based on one single counterexample. That is, one single counterexample is adequate to guide the repair. The intuition is that after
rounds of input partitioning, we typically end up repairing a small region each time. Within the small region, counterexamples closely resemble each other, and as a result, once a counterexample is repaired, others are often repaired as well. This is evidenced empirically in our experiments.

The key to design the loss function $loss(ct, N)$ is that the loss of an input $ct$ should measure how far $ct$ is from being
satisfied. 
The general idea is to adapt existing established loss functions such as cross-entropy loss for
classification tasks and mean squared errors for regression tasks. In this work, we focus on classification tasks and thus a
general form of loss functions is as follows.
\begin{align}
\label{fm:loss}
loss(ct, \hat{N}) = \sum_{i \in S}\frac{ sign(i) \cdot exp(y_i)}{\sum_{i\in L} exp(y_i)}
\end{align}
where $S\subseteq L$ is the set of desired labels (usually a singleton set) specified by the output constraint (i.e., $\omega$) and $L$ is the set of all labels in
classification; $y_i$ where $i \in L$ is the score of label $i$ given $ct$ and $\hat{N}$; and $sign(i)$ for a label $i \in S$ is
defined as follows.
\begin{align}
\label{fm:xi}
sign(i)= \begin{cases} 
			       1, & \text{if the score of label $i$ should be smaller}  \\
			       -1,              & \text{otherwise}
	     \end{cases}
\end{align}
Intuitively, $sign(i)$ is 1 for the undesired label and -1 for the other labels. The idea is thus to use the loss value as a
guideline to search for a repaired neural network $\hat{N}$ such that $\hat{N}$ produces the desired output, i.e., $\hat{N}(ct)
\vDash \omega$. 

For example, in terms of the property shown in Table~\ref{tab:exp1}, the output
constraint requires that the score of the first label is not the maximum. Let
$L=\{a,b,c,d,e\}$ be the corresponding labels of the output, i.e., $a$ is the
first label corresponding to the first dimension of the output. Then we have $S
=\{a\}$ and $sign(a)=1$ according to Formula~\ref{fm:xi}. Consequently, the
loss function is defined as follows.

\begin{align}
loss = \frac{exp(y_a)}{\sum_{i\in{\{a,b,c,d,e\}}}exp(y_i)}
\end{align}

\begin{algorithm}[t]
\caption{$repair(N, \phi, \omega, ct, \alpha, \beta, \eta)$}
\label{alg:repair}
    let $\Upsilon$ be a dictionary recording the number of times each neuron has been modified, which is initialised as empty\;
    \While{$size(\Upsilon) < \alpha$ or $timeout$}{
        $G \leftarrow$ compute the gradient of each neuron w.r.t.~$ct$ and the loss function\; 
        let $o$ be the neuron returned by $select\_neuron(G,\Upsilon,\beta)$\; 
        \If{$o$ is None}{
            \Return None\;
        } 
        let $\nabla$ be the gradient of neuron $o$, and $\zeta$ be the output of $o$ w.r.t.~$ct$\; 
        modify $N$ by setting $o$'s output as $\zeta-\eta \cdot \nabla$\;
        \If{$N(ct) \vDash \omega$ and $N$ satisfies the property}{
            \Return $N$\;
        } 
        \If{$o$ not in $\Upsilon$}{
            $\Upsilon[o]$=1\;
        }
        \Else{
            $\Upsilon[o]$=$\Upsilon[o]$+1\;
        } 
    } 
\Return None\; 
\end{algorithm}

\begin{algorithm}[t]
\caption{$select\_neuron(G,\Upsilon,\beta)$}
\label{alg:select}
$\Gamma \leftarrow$ sort the neurons according to the gradients $G$ in a descending order\; 
\For{each neuron $o$ in $\Gamma$}{
    \If{$\Upsilon[o] < \beta$}{
        \Return{o}\;
    }
} 
\Return{None}\;
\end{algorithm}

\subsection{Solving the Optimization Problem} 
We adopt a greedy strategy to solve the the optimization problem defined
by~\ref{fm:repair}. Algorithm~\ref{alg:repair} shows the details on how the
neural network is repaired. First, we initialize an empty dictionary $\Upsilon$
at line 1 to record which neurons have been modified and the number of times
they have been modified. From line 2 to line 14, we iteratively modify the
neurons in $N$. During each iteration, we first compute the gradient of each
neuron with respect to $ct$ and the value of the loss function as described
in~\ref{fm:loss}. Afterwards, we select the neuron which is most `responsible'
for the loss by invoking Algorithm~\ref{alg:select} at line 4. When a neuron is
identified, we obtain its gradient and output with respect to $ct$ at line 7.
Afterwards, we modify the neuron by tuning its output according to the gradient
at line 8. That is, like in the case of stochastic gradient descent~\cite{sgd},
we alter the neuron's output towards the opposite direction of its gradient to
decrease the violation loss. Note that we tune the neuron's output rather than
its weights to repair the neural network for the sake of efficiency. Next, we
check if the modified model is repaired at line 9, i.e., whether the modified N
satisfies the property within the input region, using the static verifier. Note
that line 9 first checks whether the counterexample has been eliminated, which
is logically speaking redundant. In practice, it serves as an efficient sanity
check and helps to reduce the number of times the verifier is called. Once a fix
is found, we return the fixed model at line 10. From line 11 to line 14, we
record the number of modifications on the selected neuron $o$. Concretely, we
first check if $o$ has ever been modified at line 11. If not, we record neuron
$o$ in $\Upsilon$ and initialize its number of modifications to 1 at line 12.
Otherwise, we increment its record by 1 at line 14.

Note that when more than one repair is performed, i.e., the loop (line 2 to line
14) executed multiple times, subsequent repairs typically do not undo the
earlier ones. There are two cases. In the first case, the repairs are for
disjoint regions of inputs, and thus the repairs are by definition independent
from each other. In the second case, if multiple repairs take place in the same
input region, because we take one counterexample from the region and solve the
optimization according to~\ref{fm:repair} during each repair, the loss typically
reduces throughout for the optimization for each repair as well as cross different
repairs.

Algorithm~\ref{alg:repair} terminates when one of the three criteria is met. 
\begin{itemize}
    \item We reach the threshold of neurons we are allowed to modify at line 2.
    \item No qualified neuron is returned or timeout at line 5.
    \item The model is repaired successfully at line 9.
\end{itemize}
Algorithm~\ref{alg:repair} either returns a repaired model which is guaranteed to satisfy the property (at line 10) or
returns \textit{None} at line 6 or at line 15 when the repair fails.

Algorithm~\ref{alg:select} shows the details about how a neuron is selected. In
Algorithm~\ref{alg:select}, we first sort all neurons of network $N$ according
to their gradients. Intuitively, the bigger the magnitude of a neuron's gradient
is, the more likely modifying the corresponding neuron would repair the model.
Note that a neuron may be selected many times. To avoid the scenario that the optimization is
stuck with a single neuron (because a single neuron can have limited influence on
the loss), we limit the number of times that a neuron can be selected to no
more than $\beta$ times at line 3.  

We remark that the greedy strategy cannot guarantee that the number of modified neurons is always the minimal. Our optimization method is
derived from the standard gradient descent method, which potentially suffers from the problem of local optima. The overall
complexity of our approach depends on the complexity of the verification algorithm and thus we evaluate it empirically in the next
section. Overall, because DeepPoly relies on abstraction interpretation techniques, it is always rather efficient.

\section{Evaluation}
\label{sec:exp}
We have implemented our approach as a self-contained toolkit called
\textsc{nRepair} with about 3k source lines of code based on PyTorch,
ONNX\footnote{https://onnx.ai/} and the state-of-the-art analyzer
ERAN\footnote{https://github.com/eth-sri/eran}. The \textsc{nRepair} toolkit as
well as all the data used in our experiments is available
online~\footnote{omitted for anonymity}.  In the following, we evaluate our approach to answer four
research questions (RQs). 
\subsection{Experimental Setup}
Our experiment subjects include 42 feed-forward networks for two kinds of two tasks, i.e., aircraft collision avoidance and image
classification.  The details are shown as follows.
\begin{itemize}
\item \emph{ACAS Xu}. ACAS Xu~\cite{acasxu} is an aircraft collision avoidance system developed for unmanned aircraft. This system
issues appropriate navigation actions to avoid collision with an intruder aircraft based on dynamic programming, which results in
a large numeric lookup table. To compress the lookup table without loss of performance, neural networks are adopted to mimic the
behaviors of the lookup table~\cite{julian2016policy}. Recently, an array of 45 DNNs was developed to reduce the lookup
time~\cite{reluplex}, which are our experiment subjects in this part of the experiment. Each of the fully connected networks has 6
hidden layers and each layer is equipped with 50 ReLU nodes. The inputs of these DNNs consists of five variables describing the
speed and relative position of both the intruder and ownership, and the outputs are the scores of five advisories. These models are subject to a set of 10 safety-critical properties~\cite{reluplex}, which we
adopt to test our approach. 

\item \emph{MNIST}. MNIST~\cite{mnist} is a dataset for image classification, which consists of 70k handwritten digits. Each digit
is represented by a $28\times28$ pixel greyscale image and the range of digits is from 0 to 9. We train three forward neural
networks over this dataset (size of training/test set is 60000/10000) and then evaluate our approach on these models. We refer the
three neural networks as FNNSmall (3 hidden layers), FNNMed (5 hidden layers) and FNNBig (7 hidden layers) respectively, and the
hidden layer size is 100 for all of the three models.  

\item \emph{CIFAR10}. CIFAR10~\cite{cifar10} is another widely used dataset. It contains 60000 $32\times32$ colour images in 10
categories. The dataset is well split, i.e., 50000 training images and 10000 test images. We train three different forward neural
networks over this dataset, and the architectures of the tree networks are same with the architectures of MNIST models described
above. Table~\ref{tab:img-models} shows the details of the image classification models used in our experiments. 

\end{itemize}

\begin{table}[t]
\caption{Neural networks of image classification}
\label{tab:img-models}
{\small \centering
\begin{tabular}{|c|c|c|c|c|}
\hline
Dataset&Model & \begin{tabular}[c]{@{}c@{}}Architecture \\ (layers$\times$ units)\end{tabular} &
\begin{tabular}[c]{@{}c@{}}Training \\ Accu.\end{tabular}& \begin{tabular}[c]{@{}c@{}}Test\\ Accu.\end{tabular}\\ \hline
 \multirow{3}{*}{MNIST}                           &FNNSmall &         $3\times100$          &0.9743         &0.9668            \\ 
                             				       &FNNMed   &          $5\times100$          &0.9803         &0.9662         \\ 
                            				       &FNNBig   &           $7\times100$          &0.9671         & 0.9386     \\ \hline
 \multirow{3}{*}{CIFAR10}             &FNNSmall &         $3\times100$          &0.6474         &0.5222           \\ 
                             			        &FNNMed   &          $5\times100$          &0.5689        &0.4966         \\ 
                                                             &FNNBig   &           $7\times100$          &0.6327         &0.4095
                                                             \\ \hline                            
     
\end{tabular}}
\end{table}

For the ACAS Xu models, among the 45 models and 10 properties, 34 models fail property 2 and 1 model fails property 7 and 1 model
fails property 8, which constitutes a total of 36 repairing tasks. 
For the MNIST and CIFAR10 models, we randomly select multiple images and verify
whether they satisfy local robustness in the $L_\infty$-norm~\cite{cw}, i.e.,
given an image $x$ with label $c$, the $L_\infty$-norm robustness is satisfied
if the model outputs label $c$ for any input in $[x-\tau, x+\tau]$. $\tau$ used
in our experiments is 0.03 for MNIST and 0.0012 for CIFAR10 (i.e.,  maximum
values in DeepPoly). We manage to identify 100 images which violate the property
for both datasets and all models except FNNBig for MNIST, for which a total of
51 images are identified. Note that for the verification of local robustness,
each image together its label is regarded as an independent property, as a
result, we total have 551 repair tasks. For each repair task, we set the timeout to 1 hour.

In the following, we report all experiment results. All experiment results are obtained on a workstation with 1 i9-9900 processor and 64GB system memory. The threshold $\alpha$ is set to 5\% of
the number of neurons in the original model, i.e., for all ACAS Xu models, no more than 15 neurons can be modified and for both
MNIST and CIFAR10, the number is 15, 25 and 35 for FNNSmall, FNNMed and FNNBig respectively. The threshold $\beta$ is set to 50
for all experiments, i.e., a neuron can be modified at most 50 times. The values of these two parameters are identified empirically. 

\subsection{Research Questions}
In this section, we aim to answer the following research questions by multiple experiments.\\

\noindent \emph{RQ1: Is \textsc{nRepair} effective at repairing neural networks?} To answer this question, we apply
\textsc{nRepair} to the 36 models of ACAS Xu, 3 models of MNIST and 3 models of CIFAR10. The step size $\eta$ used in this research
question is 0.35 for the ACAS Xu models, 0.05 for MNIST and 0.1 for CIFAR10 models. $\eta$ is the key hyperparameter in our approach
and we explain how to set the value of $\eta$ later. 

The results on repairing the ACAS Xu models are summarized in Table~\ref{tab:rq1-acasxu}, where the third
column shows the number of partitions that need to be repaired. Note that the number of partitions ranges from 2 to 22.
For each partition, we generate a counterexample (by collecting the constraints that a counterexample must satisfy, i.e., the partitioned input constraint as well as the negation of the property, and solving the constraint using mixed integer linear programming solver
Gurobi~\cite{gurobi}), and then solve the optimization problem as described in
Section~\ref{sec:app}. We then report the percentage of partitions which are
successfully repaired in the fourth column, and the number of neurons which are
modified on average to repair each partition. We observe that \textsc{nRepair}
successfully repairs 98.46\% of the partitions, which means that for almost all
of the regions, the neural network is made such that it is guaranteed to satisfy
the property. We remark that for the few regions where we fail to repair, it is
always possible to raise an alarm when an input in that region is received at
runtime. Furthermore, the number of modified neurons are kept relatively small,
i.e., 3.07 on average, or equivalently 1.02\% on average. This suggests that
only minor modifications are required to repair the neural networks. 

The results on repairing the MNIST and CIFAR10 models are summarized in
Table~\ref{tab:rq1-imgs}. Note that due to the many repair tasks (i.e., 551 in
total), we only show the average success rate and the number of modified
neurones. 
 Our approach achieves 95.72\% success rate on the two datasets on average.
Concretely, it can be observed that the success rate of most models is around 98\%, and for FNNBig of MNIST the success rate drops but still remains at a
relatively high level, i.e, 84.31\%. This suggests that our approach has the good scalability, in terms of the success
rate, and thus has the potential to be applied in practice.
 
In addition, we observe that the number of modified neurons for those models which we successfully repair is larger than those
for the ACAS Xu models, i.e., on average 7.04 and 7.13 neurons are modified. This is reasonable because for a simple model like ACAS Xu, a few
neurons can dominate a certain result, whereas for a more complex model like MNIST and CIFAR10, the classification results are typically the joint results of multiple (if not many) neurons.
We thus have the following answer to RQ1.
\begin{framed}
\noindent  \emph{Answer to RQ1: \textsc{nRepair} repairs neural network models with a high success rate and modifies a small number of neurons.}
\end{framed}

\begin{table}[t]
  \centering
  \footnotesize
  \caption{Results of repairing on ACAS Xu models}
  \label{tab:rq1-acasxu}%

    \begin{tabular}{|c|c|c|c|c|c|}
    \hline
 Property & Model & \begin{tabular}[c]{@{}c@{}}\#partitions \end{tabular} & \begin{tabular}[c]{@{}c@{}}Success\\
 rate(\%)\end{tabular}  & Avg.m & Avg.f(\%)  \\
    \hline
    \multirow{34}[68]{*}{$p_2$} & $N_{2,1}$ & 11    & 100    & 2.18  & 95.53 \\
                                & $N_{2,2}$ & 20    & 100    & 5     & 92.91 \\
                                & $N_{2,3}$ & 20    & 100    & 2.75  & 98.12 \\
                                & $N_{2,4}$ & 10    & 100    & 2.2   & 95.56 \\
                                & $N_{2,5}$ & 20    & 100    & 1.9   & 97.63 \\
                                & $N_{2,6}$ & 9     & 100    & 2.22  & 97.69 \\
                                & $N_{2,7}$ & 9     & 100    & 3.67  & 96.98 \\
                                & $N_{2,8}$ & 10    & 100    & 2.9   & 97.23 \\
                                & $N_{2,9}$ & 6     & 100    & 2.67  & 99.25 \\
                                & $N_{3,1}$ & 9     & 100    & 2.67  & 96.65 \\
                                & $N_{3,2}$ & 3     & 100    & 2     & 99.33 \\
                                & $N_{3,4}$ & 6     & 100    & 2.17  & 98.44 \\
                                & $N_{3,5}$ & 7     & 100    & 2.43  & 97.56 \\
                                & $N_{3,6}$ & 9     & 100    & 2.22  & 98.53 \\
                                & $N_{3,7}$ & 8     & 100    & 4     & 96.06 \\
                                & $N_{3,8}$ & 10    & 100    & 3.6   & 96.65 \\
                                & $N_{3,9}$ & 20    & 95     & 5.32  & 97.86 \\
                                & $N_{4,1}$ & 5     & 100    & 2.6   & 98.78 \\
                                & $N_{4,3}$ &20     & 100    & 4.05  & 95.47 \\
                                & $N_{4,4}$ & 9     & 100    & 3.22  & 96.56 \\
                                & $N_{4,5}$ & 9     & 100    & 8.11  & 97.09 \\
                                & $N_{4,6}$ & 10    & 90     & 4.11  & 98.59 \\
                                & $N_{4,7}$ & 9     & 100    & 1.22  & 98.69 \\
                                & $N_{4,8}$ & 9     & 77.78  & 3.57  & 99.07 \\
                                & $N_{4,9}$ & 4     & 100    & 4.5   & 98.61 \\
                                & $N_{5,1}$ & 22    & 100    & 3.64  & 91.67 \\
                                & $N_{5,2}$ & 20    & 100    & 2.3   & 97.34 \\
                                & $N_{5,3}$ & 2     & 100    & 2     & 99.97 \\
                                & $N_{5,4}$ & 9     & 100    & 3.56  & 96.93 \\
                                & $N_{5,5}$ & 16    & 100    & 2.25  & 98.36 \\
                                & $N_{5,6}$ & 16    & 100    & 1.44  & 98.80 \\
                                & $N_{5,7}$ & 10    & 100    & 2     & 98.24 \\
                                & $N_{5,8}$ & 9     & 100    & 2.78  & 97.43 \\
                                & $N_{5,9}$ & 9     & 100    & 1.33  & 97.97 \\
                                \hline
                      $p_7$     & $N_{1,9}$ &11     &81.82   & 4.1   & 99.38  \\
                                \hline
                      $p_8$     & $N_{2,9}$ & 11    & 100    & 3.8   & 99.33 \\
    \hline
    \multicolumn{3}{|c|}{Avg}   & 98.46    & 3.07    & 97.51\\
    \hline
    \end{tabular}%
\end{table}%

\begin{table}[t]
  \centering
  \caption{Results of repairing on image classification}
\footnotesize
    \begin{tabular}{|c|c|c|c|c|c|c|}
    \hline
    Dataset & Model & \begin{tabular}[c]{@{}c@{}}\#Repair\\ cases\end{tabular} & \begin{tabular}[c]{@{}c@{}}Success\\
    rate(\%)\end{tabular}  & Avg.m & \begin{tabular}[c]{@{}c@{}}Avg.acc \\ Local(\%)\end{tabular} & \begin{tabular}[c]{@{}c@{}}Avg.accR \\ Global(\%)\end{tabular}\\
    \hline

\multirow{4}{*}{MNIST}      &FNNSmall & 100   &100     &2.92   & 100 & 98.69 \\ \
    		                    &FNNMed   & 100   &100     &4.34   & 100 & 95.46   \\ \
                            &FNNBig   & 51    &84.31   &13.86   & 100 & 65.52  \ \\  \cline{2-7} 
                            & \multicolumn{2}{|c|}{Avg} &94.77 &7.04   & 100  &86.56 \\
    \hline

\multirow{4}{*}{CIFAR10}  &FNNSmall   & 100    & 95     &4.77    & 100   &84.98 \ \\
    				              &FNNMed     & 100    & 98     &5.71     & 100   &75.39 \\\
                          &FNNBig     & 100    & 97     &10.92   & 100   &60.42 \ \\  \cline{2-7} 
                 &\multicolumn{2}{|c|}{Avg}    &96.67    &7.13    & 100   &73.6 \\
    
 \hline    
    \end{tabular}%
   \label{tab:rq1-imgs}%
\end{table}%

\noindent \emph{RQ2: Does \textsc{nRepair}'s repair undermine the overall performance of the model?} The question asks whether the repaired model, while satisfying the property,
carries the same level of performance , i.e., whether the repaired model has an accuracy close to the original model on those inputs whose prediction results given by the original model satisfy the property. In other words, the repair should not undermine the performance of the model on those inputs which are not counterexamples to the property. 
For the ACAS Xu models, because the training or test dataset are not available, we cannot measure the accuracy
of the repaired model directly. We thus answer this question by measuring the fidelity of the repaired model with respect to the original model, which is defined as follows.
\begin{align}
Fidelity = \frac{\sum_{x\in T}\mathcal{I}(\hat{N}(x)=N(x))}{|T|}
\end{align}
where $T$ is a test set and $|T|$ is the number of samples in $T$; and
$\mathcal{I}(y)$ is an indicator function which equals 1 if $y$ holds and 0
otherwise. That is, we synthesize (based on Gaussian sampling) a test set with
10000 samples for each repair case. Algorithm~\ref{alg:gauss} shows the details
about how the test set is synthesized. The algorithm takes three parameters:
the original model $N$, the input constraint $\phi$ and output constraint
$\omega$ of a specified property. We first yield a sample $x$ (line 6-7), and
then check if the generated sample satisfies the output constraint $\omega$
(line 8). If yes, we then add it into the test set $X$. Note that we filter
those samples which fail the given property since the prediction on these
samples are supposed to be modified in order to satisfy the property. After
that, we adopt Algorithm~\ref{alg:predict} to perform the prediction with the
repaired models. The results are shown in the last column of
Table~\ref{tab:rq1-acasxu}. It can be observed that the fidelity remains across
all models, i.e., with an average of 97.51\%.

\begin{algorithm}[t]
  \caption{$synthesize\_acasxu(N, \phi, \omega)$}
  \label{alg:gauss}
  Let $(l_1, u_1), (l_2, u_2), \dots,  (l_5, u_5)$ be the lower bound and upper bound of variable $v_1, v_2, \dots, v_5$ defined by input constraint $\phi$\; 
  Let $(\mu_1, \sigma_1), (\mu_2, \sigma_2), \dots,  (\mu_5, \sigma_5)$ be the mean and standard deviation of
  variable $v_1, v_2, \dots, v_5$ respectively\; 
  Let $X=\{\}$ be the set of generated samples\;
  \While{True}{Let $x$ be an empty array with length 5.\; 
    \For{$i=1$ to $5$ }{
      $x[i] \leftarrow Gaussian(l_i, u_i, \mu_i, \sigma_i)$\;
    } 
     \If{$N(x) \vDash \omega$ }{
      $X \leftarrow X \cup \{x\}$\;
    }
    \If{$|X| == 10000$}{
      \Return{X}\;
      }
  }
  \end{algorithm}

For MNIST and CIFAR10, our goal of repairing is to make the original model
satisfy the local robustness property. That is, we repair the given model around
a set of selected images. To evaluate the the repair results on the two image
classification tasks, we report the accuracy of the model with respects to a set
of testing data containing images which are sampled within the norm, i.e., those
which satisfy the input condition. Note that all samples within the norm
should have the same label as the selected image (at the center of the norm).
The results are shown in the second last column
``Avg.acc Local" in Table~\ref{tab:rq1-imgs}. We can observe that the result is
always 100\%. This is expected as the repaired model is guaranteed to label the
images within the norm correctly.

Out of curiosity, we conduct an additional experiment as follows. We sample a set of samples throughout the input space and test the accuracy of the repaired model. Note that this is not how the repaired model is meant to be used since the repair is meant to take effect only for those inputs which satisfy the input constraint of the property. Rather, the goal is to see whether a local repair applies globally.  
Concretely, we use the test set to measure the accuracy of those repaired models relative to that of their original model, which is defined as follows.
\begin{align}
accR = \frac{\textit{accuracy of repaired model}}{\textit{accuracy of original model}}
\end{align}
The results are shown in the last column ``Avg.accR Global" of Table~\ref{tab:rq1-imgs}. It can be observed that the accuracy is high for small networks and drops
significantly as the size of the model increases. It suggests that a local repair may work globally only if the model is simple. Our interpretation is that the correlation among different neurons are complicated in large neural networks and a minor modification to some neurons may be easily magnified through the network. This suggests that a localized repair, like in our approach, is more likely to be successful in practice. We thus have the following answer to RQ2.
\begin{framed}
\noindent  \emph{Answer to RQ2: \textsc{nRepair} maintains high level of fidelity/accuracy and a local repair generated by \textsc{nRepair} does not apply globally.}
\end{framed}

\begin{figure}[t]
  \caption{Average time overhead}
  \includegraphics[width=1\columnwidth, height=0.35\textwidth]{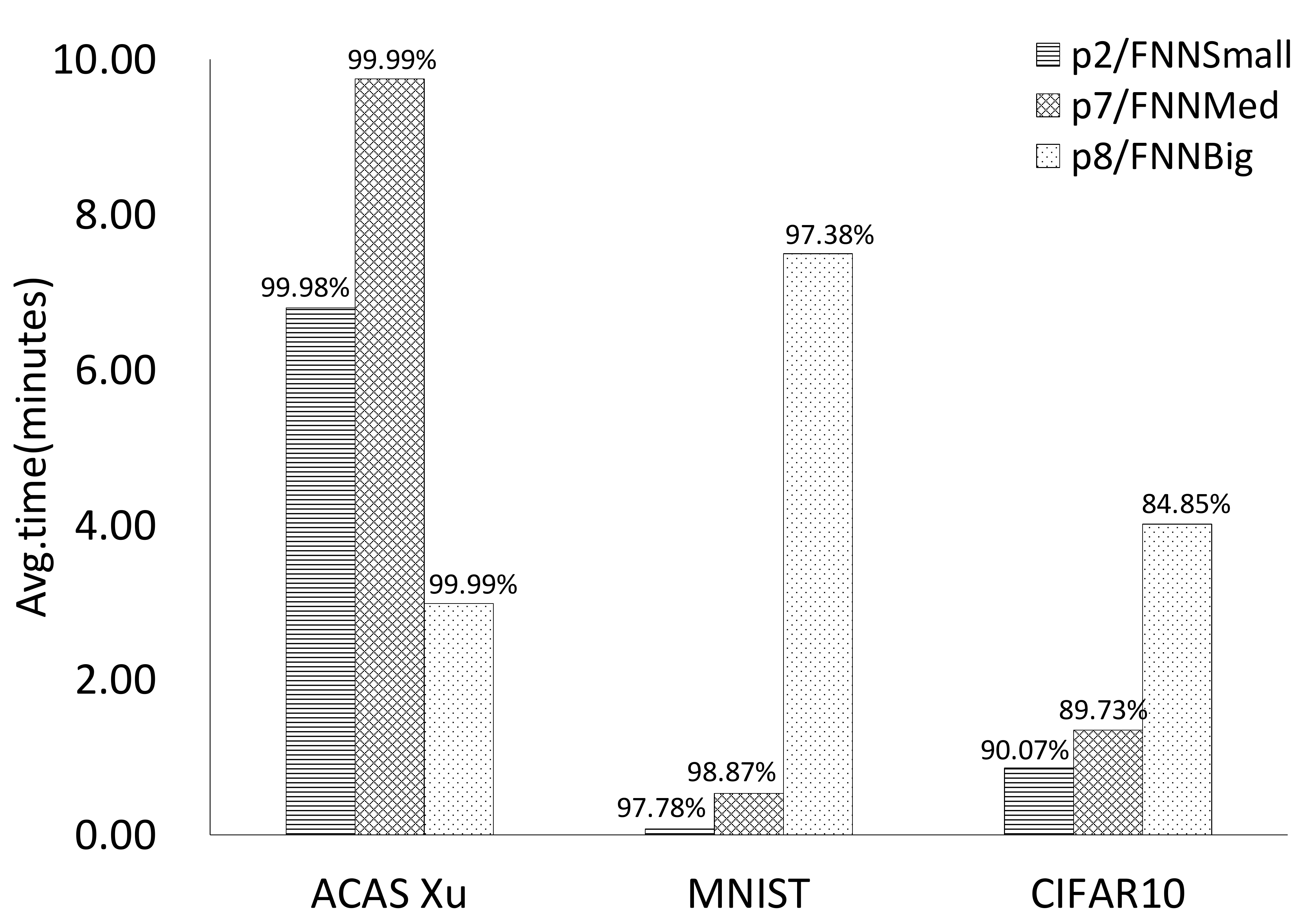}
  \label{fig:rq3}
\end{figure}

\begin{table*}[t]
  \centering
  \caption{Success rate (\%) with different $\eta$}
    \begin{tabular}{|l|l|l|l|l|l|l|l|l|l|l|l|}
    \hline
    \multirow{2}[4]{*}{Task} & \multicolumn{1}{c|}{\multirow{2}[4]{*}{Model}} & \multicolumn{10}{c|}{$\eta$} \\ \cline{3-12}
                               &               & 0.05  & 0.1   & 0.15  & 0.2   & 0.25  & 0.3   & 0.35  & 0.4   & 0.45  & 0.5 \\ 
    \hline
    \multirow{3}[6]{*}{ACAS Xu}  & $N_{3,8}$(p2)  & 80.00  & 90.00  & 100.00  & 100.00  & 100.00  & 90.00  & 100.00  & 90.00  & 100.00  & 100.00\\ \cline{2-12}
                                & $N_{1,9}$(p7)   & 63.64  & 72.73  & 81.82  & 81.82  & 81.82  & 81.82  &81.82  & 72.73  & 72.73  & 81.82  \\ \cline{2-12} 
                                & $N_{2,9}$(p8)   & 63.64  & 81.82  & 90.91  & 90.91  & 90.91  & 100.00  &100.00  & 100.00  & 100.00  & 100.00\\ \hline
    \multirow{3}[6]{*}{MNIST}   & FNNSmall        &100.00  & 100.00  & 100.00  & 100.00  & 100.00  & 100.00  & 100.00  & 100.00  & 100.00  & 100.00  \\ \cline{2-12}
                                & FNNMed          &100.00  & 100.00  & 98.00  & 97.00    & 96.00   & 97.00   & 94.00  & 96.00  & 94.00  & 91.00 \\ \cline{2-12}          
                                & FNNBig          &84.31   & 82.35   & 74.51  & 78.43    & 74.51   & 64.71   & 64.71  & 58.82  & 50.98  & 54.90 \\ \hline
    \multirow{3}[6]{*}{CIFAT10} & FNNSmall & 90.00  & 95.00  & 93.00  & 96.00  & 96.00  & 96.00  & 96.00  & 96.00  & 96.00  & 97.00\\  \cline{2-12}
                                & FNNMed   & 95.00  & 98.00  & 98.00  & 98.00  & 98.00  & 99.00  & 98.00  & 99.00  & 98.00  & 98.00  \\  \cline{2-12} 
                                & FNNBig   & 88.00  & 97.00  & 96.00  & 95.00  & 96.00  & 97.00  & 98.00  & 96.00  & 95.00  & 97.00 \\     \hline
    \end{tabular}%
  \label{tab:sr}%
\end{table*}%

\begin{figure*}[t]
  \caption{Performance of the repaired models with different $\eta$}
  \includegraphics[width=0.33\textwidth]{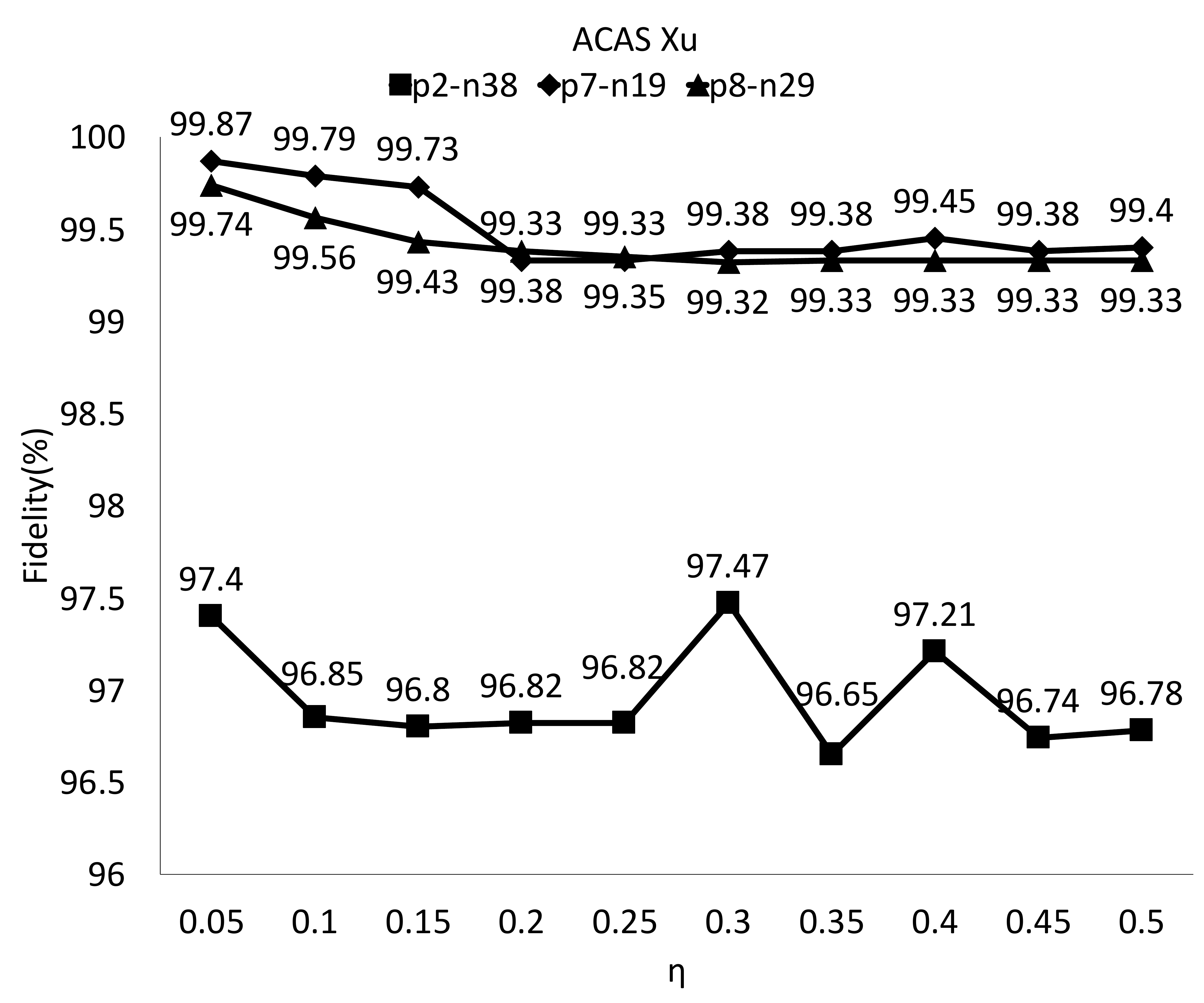}
  \includegraphics[width=0.33\textwidth]{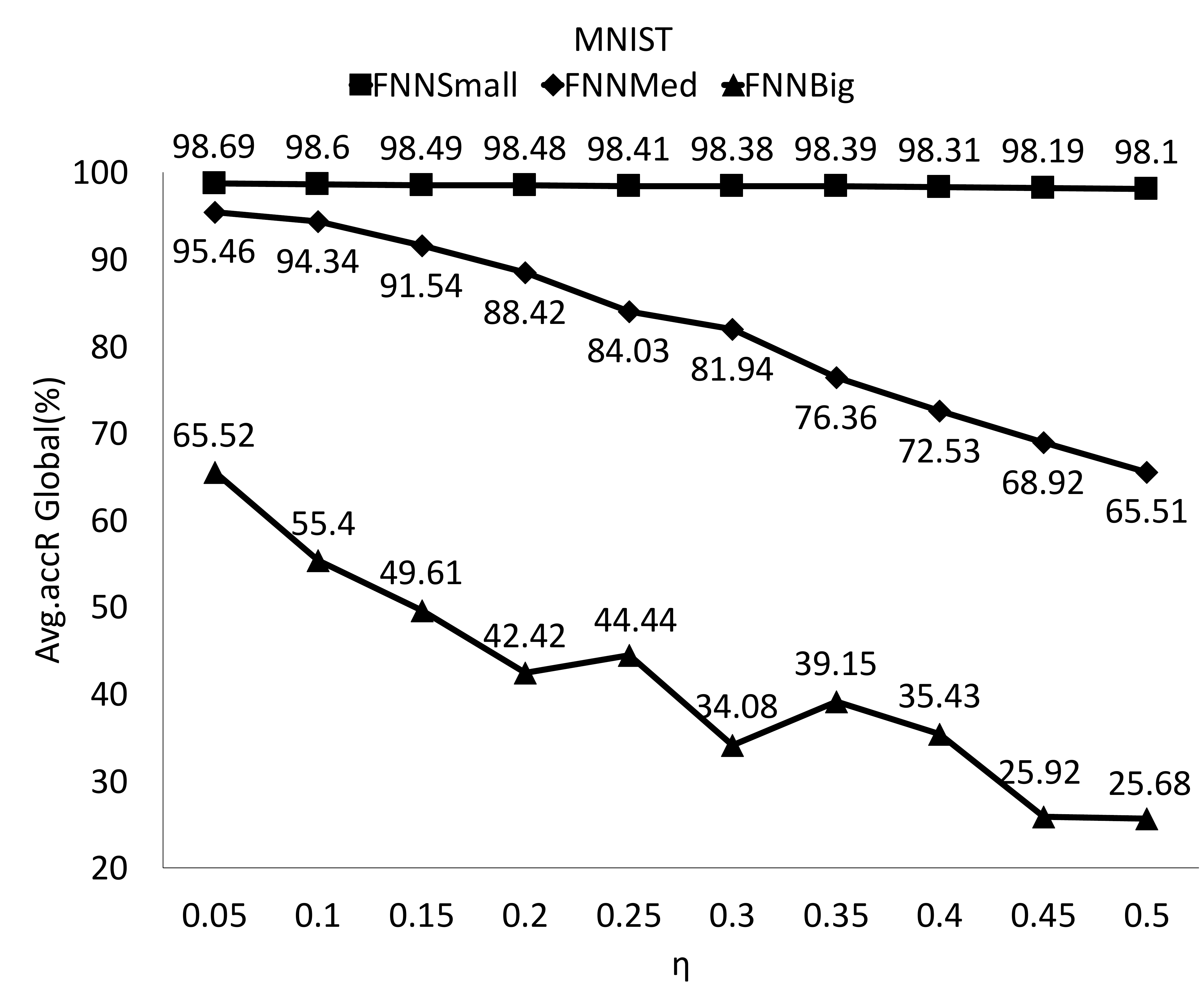}
  \includegraphics[width=0.33\textwidth]{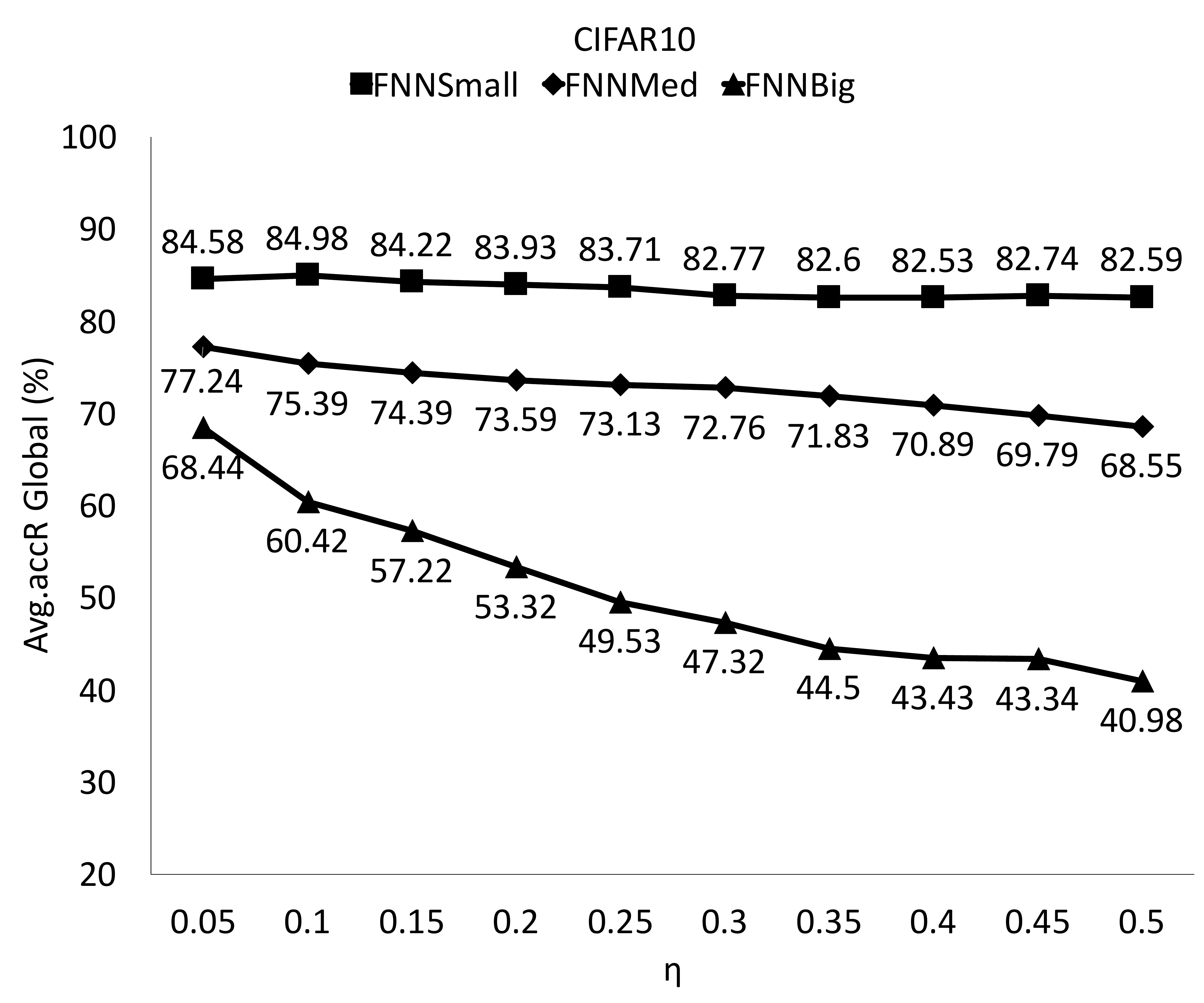}
  \label{fig:quality}
\end{figure*}

\noindent \emph{RQ3: What is the time overhead of our approach?} The most time-consuming step of our approach is the verification
part, which depends on the verification algorithm, the target model and the property. The current implement of \textsc{nRepair}
employs DeepPoly as the verifier since it is the state-of-the-art. \textsc{nRepair} could be easily refactored to take advantage
of improved neural network verifiers which we foresee will be developed in the future. 

In the following, we report the time taken
by \textsc{nRepair} to repair the models (with the same setting as in RQ1).The results are shown in Figure~\ref{fig:rq3}, where
each bar represents the average time (in minutes) of a successful repair and the percentage shown at the top of each bar is the proportion of
the verification time among the total time. For the ACAS Xu task, we show the results on three properties (i.e., property 2, 7 and 8), and for the image
classification task, we show the results on three kinds of models (i.e., FNNSmall, FNNMed and FNNBig). 

We observe that the time overhead varies across different cases. Concretely, the ACAS Xu models take the most time to repair, i.e.,
6.05 minutes on average. One possible reason is that the verified properties are more complex than the local robustness property
for MNIST and CIFAR10. In general, the more complex the property is, the more difficult it is to verify the model, and thus the
more time \textsc{nRepair} takes. This hypothesis, to some extent, can be evidenced by the time spent on repairing property 7.
According to the definition in~\cite{acasxu}, property 7 is more complex than property 2 and 8 since the verification space
(defined by the input constraint) of property 7 is the largest compared with that of property 2 and 8. We can observe that the repair
of property 7 takes the most time, i.e., 9.74 minutes. Furthermore, it can be observed that the verification time takes up 99.98\% of the repair time
for ACAS Xu, 98.01\% for MNIST and 88.22\% for CIFAR10. As expected, the verification time dominates the total execution time.
 
In general, these models are repaired within minutes, which we believe are acceptable as model training typically takes considerable time and our goal to repair a model
offline before the model is deployed. We thus have the following answer to RQ3.
\begin{framed}
\noindent Answer to RQ3: \textsc{nRepair}'s execution time depends on the underlying verifier and \textsc{nRepair} is able to repair benchmark
models within minutes. 
\end{framed}

\begin{table*}[t]
  \centering
  \caption{Average costs(minutes) with different $\eta$}
    \begin{tabular}{|l|l|l|l|l|l|l|l|l|l|l|l|}
    \hline
    \multirow{2}[4]{*}{Task} & \multicolumn{1}{c|}{\multirow{2}[4]{*}{Model}} & \multicolumn{10}{c|}{$\eta$} \\
\cline{3-12}          
                                &              & 0.05  & 0.1   & 0.15  & 0.2   & 0.25  & 0.3   & 0.35  & 0.4   & 0.45  & 0.5 \\ \hline

    \multirow{3}[6]{*}{ACAS Xu} & $N_{3,8}$(p2)  & 3.33  & 2.33  & 2.57  & 13.40  & 6.68  & 2.25  & 2.14  & 3.91  & 1.74  & 1.52   \\ \cline{2-12}          
                               & $N_{1,9}$(p7)  & 35.60  & 14.26  & 17.67  & 9.04  & 16.63  & 15.37  & 9.74  & 6.32  & 11.73  & 10.19  \\\cline{2-12}          
                               & $N_{2,9}$(p8)  & 6.77  & 6.90  & 7.31  & 4.73  & 3.30  & 4.61  & 2.98  & 2.57  & 2.49  & 22.81 \\
    \hline
    \multirow{3}[6]{*}{MNIST} & FNNSmall & 0.07  & 0.05  & 0.04  & 0.03  & 0.03  & 0.03  & 0.03  & 0.03  & 0.03  & 0.02  \\ \cline{2-12}          
                              & FNNMed  & 0.53  & 0.38  & 0.47  & 0.57  & 1.15  & 1.34  & 1.43  & 1.55  & 1.55  & 1.11 \\ \cline{2-12}          
                              & FNNBig & 7.49  & 8.65  & 7.87  & 6.74  & 6.08  & 4.56  & 6.14  & 4.20  & 1.81  & 4.45 \\ \hline
    \multirow{3}[6]{*}{CIFAR10} & FNNSmall & 1.10  & 0.85  & 0.70  & 0.65  & 0.60  & 0.50  & 0.46  & 0.43  & 0.38  & 0.36 \\ \cline{2-12}          
                                & FNNMed & 1.81  & 1.35  & 1.13  & 0.87  & 0.79  & 0.75  & 0.69  & 0.39  & 0.38  & 0.50  \\\cline{2-12}          
                                & FNNBig & 5.92  & 4.01  & 3.48  & 1.84  & 2.12  & 1.49  & 0.94  & 1.30  & 1.64  & 0.72 \\
    \hline
    \end{tabular}%
  \label{tab:avgcosts}%
\end{table*}%

\noindent \emph{RQ4: How does the value of $\eta$ influence the repair results?} At line 8 of
Algorithm~\ref{alg:repair}, we `repair' a neuron by subtracting $\eta\cdot\nabla$ from its output $\zeta$. The parameter $\eta$ has
a great impact on the repair results in many ways. Intuitively, the larger $\eta$ is, the bigger the modification is, which may have consequences on the
fidelity or accuracy as well as the success rate of the repair. Deciding the optimal value for $\eta$ is
highly non-trivial. In the following, we apply \textsc{nRepair} with different $\eta$ to understand its impact and subsequently provide practical
guidelines on how to set the value of $\eta$. 

Concretely, for the ACAS Xu models, we evaluate the influence of $\eta$ on three representative models, one for each of the three properties, i.e., model $N_{3,8}$, $N_{1,9}$ and $
N_{2,9}$ for property 2, property 7 and property 8 respectively. Note that the model for property 2, i.e., $N_{3,8}$, is selected
according to the median of the number of partitions. For MNIST and CIFAR10, we take all the three models, i.e., FNNSmall, FNNMed and FNNBig, and for each target model, to evaluate
the effect of $\eta$. For each model, we set $\eta$ to 10 different values, i.e., from 0.05 to 0.5 with a step size of 0.05, and then evaluate the
effects from three aspects, i.e., success rate, fidelity/accR and time overhead. In the following, we show the results in the
three aspects separately.

\textit{Influence on success rate}. The results of the success rate are shown in
Table~\ref{tab:sr}. It can be observed that the sensitivity to the value of
$\eta$ varies across different repair cases. Concretely, for the models of ACAS
Xu and CIFAR10, the success rate firstly increases, e.g., from 88\% to 97\% for FNNBig of CIFAR10, and then fluctuates at a high
level with the increasing of $\eta$, while for the FNNMed and FNNBig of
MNIST, the success rate drops gradually with the increasing of $\eta$. For
FNNSmall of MNIST, $\eta$ has limited impact and the success rate remains
unchanged, i.e. 100\%.

The results show that in some cases, a smaller $\eta$ may not be adequate to
effectively repair a model due to the limited modification allowed on the selected
neurons, and a bigger $\eta$ may be necessary. Furthermore, 
increasing the value of $\eta$ does not always lead to better repair since the
success rate tends to remain stable or fluctuates within a small range. Our
hypothesis is that there is often a threshold on $\eta$ such that the
magnitude of modification is adequate to ``repair'' the neuron. Imagine a case where the
model could be repaired as long as one neuron is deactivated (i.e., its output is set to be
zero). In such a case, as long as $\eta$ is large enough to reduce the
neuron's output to zero, the success rate would remain unchanged. 
In the case where the success rate drops with a larger $\eta$, an even larger $\eta$ is unlikely to work as the optimization process would probably not converge.

\textit{Influence on the performance of repaired models}. We also explore the
  impact of $\eta$ on the ``generalizability'' of the repaired models, i.e., how
  big an impact on the global accuracy the repair would lead to with different
  $\eta$. The results are shown in Figure~\ref{fig:quality}. As expected, the
  performance, i.e., fidelity or accR, decreases with the increasing of $\eta$
  in general, although the magnitude of the decline differs from model to model.
  Concretely, for small models, e.g., all models of ACAS Xu and FNNSmall of both
  MNIST and CIFAR10, are insensitive to the value of $\eta$. That is, their
  performance decreases slightly (0.59\% decline for MNIST and 1.9\% decline for
  CIFAR10). For big models, the performance of repaired results drops
  significantly when the value of $\eta$ increases, i.e., nearly 40\% declines
  for MNIST and 28\% declines for CIFAR10. This result is expected as a larger
  $\eta$ means a greater modification during each iteration in
  Algorithm~\ref{alg:repair}. Furthermore, as discussed before, deeper and
  larger networks tend to magnify small modification. 

 \textit{Influence on time overhead}. Intuitively, a larger $\eta$ may
  accelerate the repair. We show the average time of successfully repair in
  Table~\ref{tab:avgcosts}. We can observe that in most cases, the time overhead
  decreases when an increased $\eta$ value. Specially, for the FNNBig of
  CIFAR10, the time overhead decreases significantly, e.g., the repair with 0.5
  on FNNBig is 8.2 times faster than that with $0.05$. However, there is an
  exception where the costs increase, i.e., FNNMed of MNIST. This is because
  that the a large $\eta$ may lead to over modification on a single neuron and
  make the neuron ``jump over" the ``solution". In this case, more neurons will
  be involved, and thus more time is spent. We thus have the following answer to RQ4.

\begin{framed}
\noindent \emph{Answer to RQ4: In general, a small $\eta$ leads to a repaired model with a high fidelity and more time spent on repairing. Our practical guideline is thus to 
have a small $\epsilon$ as long as the model can be repaired.}
\end{framed}

\subsection{Threats to Validity}
\noindent \emph{The dependence on the verifier} We only evaluate our approach with the DeepPoly verifier. We choose DeepPoly as
it is the state-of-the-art at the time of writing. Different verifiers may lead to different performance. In fact, our approach is
orthogonal to the rapid development of neural network verification techniques. As long as the verifier used in our framework is
sound (i.e., when the verifier returns holds, the property actually holds), our approach works. \\ 

\noindent \emph{The size of repaired network} As illustrated in Algorithm~\ref{alg:overall},  our approach returns an
repaired network which is the result of assembling the repaired result of each erroneous partition. Thus, the size of the final repaired network depends on
the number of erroneous partitions. In our experiments, for the ACAS Xu DNNs, the number of partitions
varies from 2 to 20, and thus the size of the repaired network is 2 times to 20 times bigger than the original one. The blowup
could be reduced by combining common parts of the repaired network, which we will study in the future work. \\

\noindent \emph{Limited number of images} We evaluate our approach with 100 images in fixing the local robustness property. The
amount of images may not be adequate. Our experimental setting is largely adopted from existing work, i.e., Goldberger \emph{et al.}~\cite{mm} and
Singh \emph{et al.}~\cite{deeppoly}, both used 100 images for evaluation. In our experiments, we take the limited number of images because that
it is difficult to obtain qualified test cases. Images selected from the testing dataset must be correctly predicted by the
original model but fail to be verified. Meanwhile, at least one counterexample can be found when one image can not be verified.
Under these constraints, the total number of qualified images is small. For example, we only found 51 qualified
images for the FNNBig on the whole MNIST testing set (10000 images in total). However, our approach can be easily extended
to large scale datasets if more qualified images are available.





\section{Related works}
This work is closely related to existing proposals on repairing unexpected
behaviours of DNNs. Existing approaches on this topic can be roughly categorized
into three groups. The first is network patching~\cite{patch2018,
kauschke2018towards} which uses an auxiliary classifier to estimate if a patch
should be applied. The second is adversarial retraining~\cite{li2016general,
ma2018deepgauge} which firstly identifies or synthesises a group of inputs which
lead to the unexpected behaviours, and then retrain or fine-tune the neural
network with these inputs. Works in this category mainly focus on how to
efficiently generate samples for DNN repairing. For example,  Ren \emph{et al.}
propose a method named FSGMix to augment training data with the guidance of
failure examples~\cite{FSGMix}.
The last category is weights modification which directly modifies the weights of the
neurons. Approaches in this category differ in the two aspects, i.e., the
selection of neurons the weights of which need to be modified and how the new weights are computed.
Arachne~\cite{sohn} identifies the weights which are deemed related to the
specified misbehaviours and then uses a PSO algorithm to generate a patch for
these selected weights. While Goldberger \emph{et al.}~\cite{mm} directly
select the weights connected to the output layer and computes the new weights
by solving a verification problem. We remark that none of the above-mentioned methods
guarantee that the repaired network always behave correctly with respect to the properties.

The approach in~\cite{mm} is similarly verification-based. However, the repair problem they tackle is different from ours. More specifically, while we aim to repair a neural network model satisfies a 
user-specified property always, their goal is to repair a neural network such that the modified model behaves correctly on one or multiple specific inputs. That is, their approach is devised to correct the DNN’s behavior on certain concrete inputs,
and thus can not be applied to in scenarios such as ours, where the repaired model is expected
to behave correctly on all inputs satisfying the input constraint (unless the input constraint is so restrictive such as only a few concrete inputs are allowed). The limitations of their approach is
also evidenced by their attempt on the ACAS Xu models reported in~\cite{mm},
i.e., they tried to apply their algorithm to repairing the ACAS Xu models to satisfy one of the 10 properties~\cite{acasxu} but failed.


This work is related to verification of DNNs in general.
Many works in this category approximate the nonlinear activation functions as linear constraints
to facilitate constraint solving, like the MIPVerify~~\cite{MIPVerify} which is
based on mixed integer programming; Some works verify a target network by analysing the reachability layer-by-layer, such as
DeepPoly~\cite{deeppoly} which is based on abstract interpretation and Neurify~\cite{neurify} which computes the bounds of each neuron's output
based on symbolic interval analysis and linear relaxation. Satisfiability modulo theories (SMT) based techniques are also widely used for verifying DNNs. A classical SMT based verifier for deep neural
networks is Reluplex~\cite{reluplex}. Another popular SMT-based work for
verifying deep neural networks is Marabou~\cite{marabou} which answers queries
about user-provided properties by transforming the queries into constraint
satisfiability problems. Different from the above-mentioned verification techniques, we focus on repairing DNNs.


\section{Conclusion}
In this work, we propose an approach for repairing DNN based on existing verification techniques. Our approach is based on selectively modifying the activation weights for a small number of neurons so that the resultant model is guaranteed to satisfy the property. To identify the neurons which are most relevant to the violation of the property, we reduce the problem to an optimization problem by defining a loss function, and then select the relevant neurons according to the gradients of the loss on each neuron. We show that our approach effectively repairs a range of benchmark models with specific properties over the ACAS Xu, MNIST and CIFAR10 datasets. 
\clearpage
\bibliographystyle{plain}
\bibliography{ref}
\end{document}